\pdfoutput=1

\documentclass[11pt]{article}

\usepackage{EMNLP2023}
\usepackage{graphicx}
\usepackage{adjustbox}
\usepackage{booktabs}
\usepackage{multirow}
\usepackage{times}
\usepackage{latexsym}

\usepackage[T1]{fontenc}
\usepackage[utf8]{inputenc}
\usepackage{microtype}

\usepackage{inconsolata}

\title{USTHB at NADI 2023 shared task: Exploring Preprocessing and Feature Engineering Strategies for Arabic Dialect Identification}

\author{Mohamed Lichouri\\
  LCPTS-USTHB, Algiers, Algeria \\
  \texttt{mlichouri@usthb.dz}
  \And 
  Khaled Lounnas, Aicha Zitouni \\
  LCPTS-USTHB, Algiers, Algeria \\
  CRSTDLA, Algiers, Algeria \\
  \texttt{\{k.lounnas, a.zitouni\}@crstdla.dz}
  \AND
  Houda Latrache \\
  CRSTDLA, Algiers, Algeria \\
  \texttt{h.latrache@crstdla.dz}
  \And 
  Rachida Djeradi \\
  LCPTS-USTHB, Algiers, Algeria \\
  \texttt{rdjeradi@usthb.dz}\\
  }

\date{}

\begin{document}
\maketitle
\begin{abstract} 
In this paper, we conduct an in-depth analysis of several key factors influencing the performance of Arabic Dialect Identification NADI'2023, with a specific focus on the first subtask involving country-level dialect identification. Our investigation encompasses the effects of surface preprocessing, morphological preprocessing, FastText vector model, and the weighted concatenation of TF-IDF features. For classification purposes, we employ the Linear Support Vector Classification (LSVC) model. During the evaluation phase, our system demonstrates noteworthy results, achieving an F1 score of 62.51\%. This achievement closely aligns with the average F1 scores attained by other systems submitted for the first subtask, which stands at 72.91\%. 
\end{abstract}

\section{Introduction}
\label{intro}

In recent times, the exploration of Arabic dialects has emerged as a focal point for numerous dedicated researchers, with various workshops dedicated to this captivating domain. Over the years, several pioneering studies have made substantial contributions to the task of Arabic language identification, particularly in areas closely aligned with the experiments conducted in this study. Examples of these influential contributions include MADAR 2019, as presented in Bouamor et al.'s work \cite{bouamor2019madar}, the insights gained from NADI 2020 \cite{mageed2020nadi}, and the valuable research findings unveiled in NADI 2021 \cite{abdul-mageed-etal-2021-nadi} and NADI 2022 \cite{abdul-mageed-etal-2022-nadi}.

In the realm of Arabic text preprocessing, previous research has delved into techniques like surface cleaning and morphological analysis. For instance, the work of Ayedh et al. \cite{ayedh2016effect} where they investigated the effect of preprocessing tasks on the efficiency of the Arabic Document classification system. 

Furthermore, research by Bouamor et al. \cite{pasha-etal-2014-madamira} is instrumental in the development of preprocessing tools such as MADAMIRA, which has been valuable in Arabic text analysis and dialect identification. Their work has served as a foundation for subsequent research in this domain and has implications for feature extraction, a crucial part of our experiments.

In the context of feature extraction, recent advancements have been made using the FastText model for Arabic text analysis. Notably, the work of Kaibi et al. \cite{kaibi2019compare} introduced innovative approaches that leverage FastText for efficient feature extraction in Arabic text classification tasks. Their methods have inspired our approach in Experiment 3.

Moreover, the weighted union of features has gained attention as an effective strategy for improving classification performance. Research by Lichouri et al. \cite{lichouri-etal-2021-arabic} demonstrated the benefits of weighted feature fusion in multilingual text classification, highlighting its adaptability to diverse linguistic datasets. This approach aligns closely with our Experiment 4, which focuses on the weighted concatenation of TF-IDF features.

These earlier studies, along with the recent advancements in FastText-based feature extraction and weighted feature fusion, have collectively paved the way for new idea, providing insights and methodologies that are closely related to the experiments conducted in this paper. They reflect the growing importance of Arabic language identification in the context of diverse linguistic data sources and dialectal variations.

This paper, rather than introducing innovative solutions or groundbreaking insights for NADI 2023 \cite{abdul-mageed-etal-2023-nadi}, serves as a concise consolidation of existing knowledge. It is structured as follows: Section \ref{data} provides an overview of the dataset used in our study. In Section \ref{aprch1}, we present our proposed system, encompassing detailed descriptions of the applied data cleaning steps and preprocessing methods, which are further elaborated on in Section \ref{prep}. Section \ref{feat} delves into the process of feature engineering, and Section \ref{res} is dedicated to discussing the findings and their significance. Finally, our paper concludes in Section \ref{conc}, summarizing the key takeaways and contributions.

\begin{table*}[!t]
\centering
\begin{tabular}{l|c|c|c|c|}

\cline{2-5}
 & \multicolumn{4}{c|}{\textbf{Closed Country-level Dialect Identification subset}}  \\ \cline{2-5} 
 & \textbf{Train} & \textbf{Dev} & \textbf{Test} & \textbf{Total} \\ \hline
\multicolumn{1}{|l|}{\textbf{\# sentences}} & 18,000 & 1800 & 3600 & 23400 \\ \hline
\multicolumn{1}{|l|}{\textbf{\# words}} & 260k & 26.1k & 52.1k & 338.2 k  \\ \hline
\multicolumn{1}{|l|}{\textbf{Max \# word per sentence}} & 59 & 55 &  57 & -  \\ \hline
\multicolumn{1}{|l|}{\textbf{Min \# word per sentence}} & 2 & 3 & 3 & - \\ \hline
\multicolumn{1}{|l|}{\textbf{Max \# char per sentence}} & 296 & 275 & 284 & -  \\ \hline
\multicolumn{1}{|l|}{\textbf{Min \# char per sentence}} & 9 & 12 & 11 & -  \\ \hline
\end{tabular}
\caption{Dataset statistics after applying some pre-processing steps}
\label{tab1}
\end{table*}

\section{Description of the Dataset}
\label{data}
Nuanced dialect identification at the country level (NADI) involves the sophisticated analysis and classification of regional language variations within a specific country. It goes beyond mere recognition of a country's official language and aims to pinpoint the distinctive linguistic nuances that exist within different regions or communities. This process typically employs advanced computational and linguistic techniques to identify, categorize, and differentiate these dialects based on phonological, morphological, syntactical, and lexical features. 

To analyze the provided text samples and determine which dialects are the most confusing and the most distinguished, let's take a closer look at the text content and the labeled dialects.  the clarity of the dialects in the given text. Here's an analysis:
\begin{itemize}
    \item \textbf{Most Confused Dialects}:
    \begin{enumerate}
        \item \textbf{Palestine and Jordan}:  Both Palestine and Jordan share similarities in their dialects due to geographic proximity. This similarity might lead to confusion between these two dialects in certain contexts.
        \item  \textbf{Saudi Arabia and Oman}:  Both Saudi Arabia and Oman share some linguistic similarities, and confusion might arise when common Arabic terms are used.
        \item  \textbf{Syria and Lebanon}:  Syria and Lebanon also share linguistic and cultural similarities, making it possible for their dialects to be confused, especially in written text.
    \end{enumerate}
\item \textbf{Most Distinguished Dialects}:
    \begin{enumerate}
        \item \textbf{North\_Africa country}:
        The presence of a hashtag that unmistakably signifies the dialect as one of the North African varieties, like Algerian, is evident in the text. The inclusion of the term 
        "Jazā'irīyah"  (Algerian) within the text serves as a distinguishing factor. Moreover, references to Tunisian politics are made through hashtags such as "\# tūnis" and "\#Tunisie." The overall context and these hashtags unequivocally establish a connection to Tunisia. Additionally, the phenomenon of code-switching, which occurs between Algerian, Libyan, Moroccan and Tunisian dialects, and the French language, is specifically linked to these dialects.
    \end{enumerate} 
\end{itemize}

These three subsets (Closed Country-level Dialect Identification, Closed Dialect to MSA MT, and Open Dialect to MSA MT) include a total of 23.4K tweets from 18 Arab countries. We divided these subsets into three parts: train, dev, and test, for which we addressed some statistics in Table \ref{tab1}, after applying a number of pre-processing steps.As can be noticed from Table \ref{tab1}, the minimum number of words and characters per sentence in the training and development sets is 2 and 9 respectively.

\begin{figure*}[!h]
    \centering
    \includegraphics[width=0.89\textwidth, height=10cm]{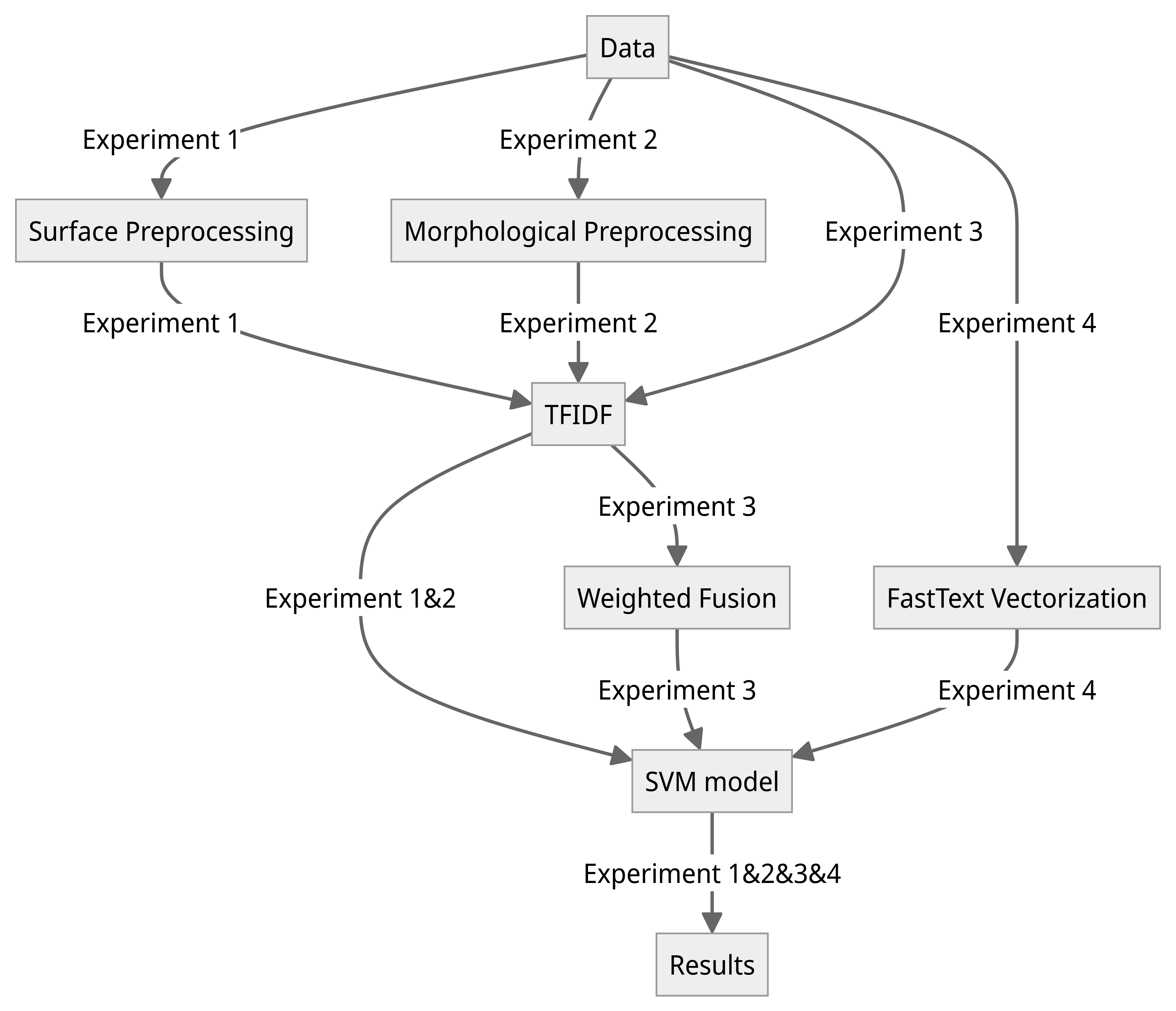}
    \caption{Proposed system for Closed Arabic Dialect Identification.}
    \label{fig1}
\end{figure*}

\section{Proposed system} 
\label{aprch1}

\subsection{Data Preprocessing} 
\label{prep}
As we delve into the task of identifying the dialect of Twitter messages, it becomes imperative to accurately capture all the necessary information while effectively removing undesired elements. To achieve this, we have adopted a two-phase preprocessing approach:

\begin{description}
    \item[Phase 1: Surface Preprocessing] In this initial phase, we execute a series of essential steps. These include Arabic Letter Normalization, removal of punctuation and emojis, removal of stop words, diacritics elimination, and the removal of non-Arabic content. These processes collectively ensure that the text remains clear, consistent, and free from distractions \cite{krouska2016effect}.
    \item[Phase 2: Morphological Preprocessing] In the subsequent phase, our focus shifts to linguistic intricacies. Here, we employ lemmatization and stemming to simplify word forms, thereby assisting in the identification of core word meanings and structures \cite{pradana2019effect}.
\end{description}
    
Throughout both phases, we change and fine-tune various techniques to arrive at the most suitable configuration for our preprocessing pipeline.

\subsection{Feature engineering} 
\label{feat}
Our system is structured into four distinct phases that can be applied individually or in tandem (see Figure \ref{fig1}). The initial two phases, Surface Preprocessing and Morphological Preprocessing, have been elaborated upon in the preceding section. The subsequent phases are as follows:

\begin{description}
    \item[Phase 3: Feature Extraction] In this stage, we employ two models. Firstly, the FastText \cite{alessa2018text} model undergoes training in two modes: supervised and unsupervised, utilizing the training dataset. Then, we use this model to extract features from the development and test datasets. Secondly, we employ the TF-IDFVectorizer, which employs three analyzers (Word, Char, and Char\_wb), each with varying n-gram ranges. In the default configuration, we combine these three features, assigning equal weights of 1 to each.
    
    \item[Phase 4: Weighted Union] In this phase, we combine the three TF-IDF features using a weight vector comprising three distinct values (w1, w2, w3) corresponding to the Word, Char, and Char\_wb analyzer, respectively \cite{lichouri-etal-2021-arabic}.
\end{description}

\begin{table}[!h]
\caption{The various combinations and parameter used in our work}
\label{houda}
\begin{tabular}{|c|c|}
\hline
\textbf{Settings}   & \textbf{Range}        \\ \hline
ngram\_range        & \begin{tabular}[c]{@{}l@{}}(m,n) with m=1 to 3 \\ and n=1 to 10 \end{tabular} \\ \hline
tfidf\_weights      & 0.5 - 1               \\ \hline
tfidf max\_features & 1000 -25000           \\ \hline
SVM                 & C=100, gamma=1-10  \\ \hline
fasttext\_supervised  & epoch=100, loss='ova' \\ \hline
fasttext\_unsupervised  & \begin{tabular}[c]{@{}l@{}} epoch=100, ws=6\\ model='skipgram'\\ dim=1000\end{tabular}\\ \hline
\end{tabular}
\end{table}

After defining the four distinct phases, we opted to work with four designed experiments based on our previous works \cite{lichouri2018word, abbas2019st, lichouri2020simple}, each featuring unique configurations:

\textbf{Experiment 1 \cite{lichouri-etal-2021-preprocessing, lichouri2020speechtrans}:} In this experiment, we initiated with the first phase, by exploring all possible combinations of surface processing techniques. Then, we proceeded to the third phase, employing a straightforward union of TF-IDF features. During feature extraction, we varied the n-gram values (ranging from $n=1$ to $10$). Finally, the SVC classifier was trained.

\textbf{Experiment 2 \cite{lichouri2020profiling}:} In this scenario, we commenced with the second phase, by examining various combinations of morphological processing techniques. Similar to Experiment 1, we followed this with the third phase, involving the union of TF-IDF features with n-gram variation. The SVC classifier was subsequently trained.

\textbf{Experiment 3:} For this experiment, we exclusively ran the third phase, employing the FastText model for feature extraction. Following this, the SVC classifier was trained.

\textbf{Experiment 4 \cite{lichouri-etal-2021-arabic}:} In this instance, we solely executed the fourth phase, applying a weighted union of TF-IDF features for feature extraction. The SVC classifier was then trained.

After conducting these four experiments on both the training and development datasets, we carefully recorded the best results achieved for each experiment, along with their optimal configurations, which are presented in Table \ref{houda}.

\begin{table}[h!]
\centering
\begin{tabular}{|l|c|c|c|c|}
\toprule
\textbf{Runs/Exp} & \textbf{Exp\_1} & \textbf{Exp\_2} & \textbf{Exp\_3} & \textbf{Exp\_4} \\
\midrule
Run 1 & 54.42 & 59.92 & 55.77  & \textbf{60.25} \\
Run 2 & 53.1 & \textbf{60.02} & \textbf{61.24} & 60.83 \\
Run 3 & 53.96 & 59.56 & 60.66 & \textbf{62.84} \\
\bottomrule
\end{tabular}
\caption{Obtained F1 Scores using the proposed system. Where: Exp\_{i} design either experiment 1, 2, 3 or 4. }
\label{houda1}
\end{table}

\subsection{Obtained results}
\label{res}
As part of our participation in the Closed Country-level Dialect ID project, we conducted various experiments aimed at enhancing the performance of our Support Vector Classifier (SVC) model. These experiments involved exploring different preprocessing steps, including Surface Preprocessing (SP) and Morphological Preprocessing (MP), the utilization of various vectorization models such as supervised and unsupervised fastText, as well as TF-IDF, and the incorporation of Weighted Fusion (WF).

In order to optimize our model, we fine-tuned several crucial configuration parameters, with a primary focus on the selection of n-grams and limiting the maximum number of most frequent terms used in TF-IDF.

The results of our experiments (see Table \ref{houda1}) revealed that the supervised FastText vectorization model outperformed the others by achieving the highest F1 score, reaching 62.84\% (Experiment 4). It was closely followed by TF-IDF with an F1 score of 61.24\%, while unsupervised FastText exhibited slightly lower performance with an F1 score of 60.25\%.

However, an intriguing observation occurred when we introduced preprocessing (SP and MP) into the process. To our surprise, the F1 score experienced a significant drop, reaching only 53.10\% compared to using vectorization alone (TF-IDF). This decline in performance can be attributed to potential information loss, alterations in word forms through lemmatization, as well as the removal of punctuation. These transformations may have disrupted class balance, thus negatively impacting the model's ability to perform accurate classification. 

\section{Conclusion} 
\label{conc}
In conclusion, our rigorous experimentation and analysis in the realm of Arabic Dialect Identification, particularly within the context of country-level dialect identification, have yielded valuable insights into the effectiveness of various approaches. Through multiple testing and comparison, we have identified that the optimal strategy for achieving high accuracy in this task involves the weighted concatenation of TF-IDF features followed by the application of the FastText model.

Our findings underscore the significance of feature engineering and the fusion of weighted TF-IDF matrices as a robust foundation for this classification task. Furthermore, the integration of FastText, with its capacity to capture semantic relationships and nuances, enhances the overall accuracy and effectiveness of our dialect identification system.

The remarkable alignment between our system's performance, yielding an F$_1$ score of 62.51\% for closed country-level dialect identification, and the average scores achieved by other submitted systems (72.91\% for the first subtask), underscores the competitiveness and efficacy of our approach.

\bibliography{anthology,custom}
\bibliographystyle{acl_natbib}

\end{document}